\documentclass[conference,a4paper,twocolumn]{IEEEtran}

\IEEEoverridecommandlockouts

\ifCLASSINFOpdf
\else
\fi
  \usepackage{tipa}
  \usepackage[pdftex]{graphicx}
  \graphicspath{{./figures/}}
	\usepackage{epsfig}
	\usepackage{epstopdf}

\usepackage[cmex10]{amsmath}
\usepackage{algorithmic}

\usepackage{array}

\ifCLASSOPTIONcompsoc
  \usepackage[caption=false,font=normalsize,labelfont=sf,textfont=sf]{subfig}
\else
  \usepackage[caption=false,font=footnotesize]{subfig}
\fi
\usepackage{url}

\begin{document}



\title{Temporal aggregation of audio-visual modalities for emotion recognition}


\author{
\IEEEauthorblockN{
Andreea Birhala$^{*}$, Catalin Nicolae Ristea$^{*}$, Anamaria Radoi$^{*}$, and Liviu Cristian Dutu$^{{\dag}}$
}
\IEEEauthorblockA{
$^{*}$Research Center CAMPUS, University Politehnica of Bucharest, Bucharest, Romania\\
$^{\dag}$Xperi Corporation, Bucharest, Romania\\
Email: anamaria.radoi@upb.ro}
\thanks{This work has been partially supported by the Ministry of Innovation and Research, UEFISCDI, project SPIA-VA, agreement 2SOL/2017, grant PN-III-P2-2.1-SOL-2016-02-0002.}}


%


\maketitle


\begin{abstract}
Emotion recognition has a pivotal role in affective computing and in human-computer interaction. The current technological developments lead to increased possibilities of collecting data about the emotional state of a person. In general, human perception regarding the emotion transmitted by a subject is based on vocal and visual information collected in the first seconds of interaction with the subject. As a consequence, the integration of verbal (i.e., speech) and non-verbal (i.e., image) information seems to be the preferred choice in most of the current approaches towards emotion recognition. In this paper, we propose a multimodal fusion technique for emotion recognition based on combining audio-visual modalities from a temporal window with different temporal offsets for each modality. We show that our proposed method outperforms other methods from the literature and human accuracy rating. The experiments are conducted over the open-access multimodal dataset CREMA-D.

\end{abstract}



\IEEEpeerreviewmaketitle


\section{Introduction}
 Automatic detection of human emotions has become an important area of research due to the technological development that occurred in the human-computer interaction domain (e.g., social robots \cite{cavallo2018emotion, 8906635}, monitoring systems for car drivers' condition \cite{4490039}). In order to increase the accuracy of emotion recognition systems, most of the currently developed methods incorporate multimodal information (e.g., facial and speech features) \cite{ristea2019emotion, 10.1145/2512530.2512533, beard-etal-2018-multi, 8925444}. Facial expressions represent one of the most important modes of communication through which people express their emotions and intentions. In addition to facial expressions, people also express their feelings through speech; e.g, speech inflection, vocal intensity are characteristics that contain information about the emotional state of a subject. 

Each person is unique and can express emotions in their own characteristic way, depending on their culture, age, gender or previous life experiences \cite{8070966}. Nevertheless, there are common characteristics that can be exploited in order to obtain an accurate classification system. In general, most recognition systems consider only 6 types of emotions (e.g., anger, happiness, surprise, disgust, contempt, anxiety) \cite{ekman1992argument}. According to the Facial Action Coding System (FACS), each human emotion can be described through a combination of several Facial Action Units (FAUs) \cite{908962}. More precisely, the FACS refer to a combined set of facial muscle movements that correspond to a displayed emotion. The basic element in this coding system is Action Unit (AU) and each AU is related to the contraction of one or more facial muscles.

Emotion recognition systems using only visual information (i.e., video frames) can be mainly classified into static and dynamic methods depending on the feature representations. In static-based methods, the features are encoded with spatial information from singular frames without taking into consideration the temporal extent, whilst dynamic-based methods consider the temporal relation between continuous frames from the input sequence. In the case of static-based methods, state-of-the-art deep neural networks architectures (e.g., VGG \cite{simonyan2014very}, ResNet \cite{he2016deep}) have been proposed for feature extraction, whilst the classification into emotion categories is performed using a Support Vector Machine (SVM) module \cite{method2}.

Due to the increased interest in developing real-world scenarios datasets and also the increased computer processing capabilities, recent approaches are based on deep learning techniques that are able to extract both facial and audio discriminant information. In a recent paper \cite{ristea2019emotion}, we have shown that not very deep convolutional neural network (CNN) architectures are able to extract meaningful information regarding emotion categories from both video frames and spectrograms of audio signals. By combining the audio-video information, we managed to achieve an increase of almost 7 \% compared to the case when only video data is considered.



Considering the behavioral differences between people and the diverse modes of communicating their feelings, methods addressing person-specific affective understanding have been also developed. In \cite{affectivememory}, Grow-When-Required Networks and personalized affective memories are used to learn individualized aspects of emotions. However, the complexity of the proposed model limits the real-time usage of the proposed solution.

In order to include temporal dynamic characteristics between video frames, Beard et. al proposed a recurrent multi-attention (RMA) mechanism with shared external memory that is updated over multiple iterations of analysis \cite{beard-etal-2018-multi}. This approach allows relevant memories to persists over multiple hops. The method achieved a maximum accuracy of 65 \% on the CREMA-D dataset, comparable to the human rating accuracy reported for this dataset (i.e., 63.6 \% \cite{cao2014crema}). 

In an attempt to exploit the complementary information brought by diverse modalities (i.e., audio and video), a Multimodal Emotion Recognition Metric Learning (MERML) was defined in \cite{8935376}. The learned metric was further used by SVM with Radial Basis Function (RBF) kernel. 


In this paper, we propose a novel multimodal architecture that combines visual and audio features extracted from random selections of analysis windows within individual temporal segments of the input video. Thus, the temporal aggregation of audio and video allows for asynchronous inputs of the two considered modalities. We tested our solution on the CREMA-D \cite{cao2014crema}, a widely used audio-visual dataset in the multimodal emotion recognition field.

The rest of the paper is organized as follows. Section II introduces the multimodal model architecture. Section III describes the dataset used for experiments, whereas section IV presents the experimental results. Finally, section V concludes the paper.

\section{Proposed Approach}

\begin{figure}[!t]
  \centering
  \includegraphics[width=0.45\textwidth]{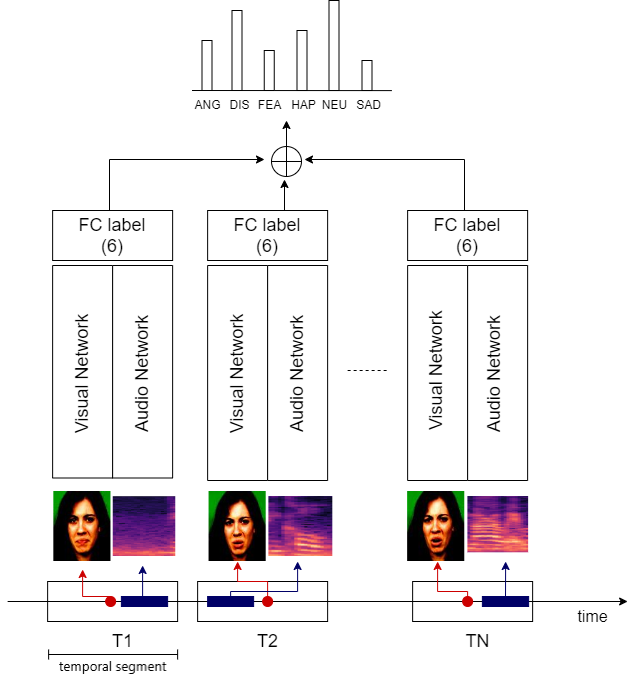}
  \caption[]{Our proposed temporal aggregation mechanism. FC label is a 6-dimensional vector of class probabilities obtained after passing the audio-visual features through the Fully Connected (FC) layer.}
  \label{fig:tbn}
\end{figure}

Inspired by the solution for action recognition presented in \cite{kazakos2019epic} and by how the human brain processes audio-visual information, we propose a temporal aggregation mechanism in order to combine modalities within a range of temporal offsets. In this mechanism we explore the fusion between audio and visual inputs within a temporal window, that will allow the model to be trained with asynchronous inputs from both modalities. Our proposed temporal aggregation mechanism is shown in Fig~\ref{fig:tbn}. In the following, we denote by $r_v$ the sampling rate of the video sequence and by $r_a$ (different from $r_v$) the sampling rate of the audio signal.


The input video sequence is divided into $N$ temporal segments of equal length. For each temporal segment, we randomly select a video frame $j$ and we randomly choose the center $c$ of the audio signal window between $[j/r_v - b, j/r_v + b]$ seconds. The audio signal used in the analysis of the current temporal segment is considered between $[c - d/2, c + d/2]$ seconds (i.e., samples between $\lceil r_a(c-d/2) \rceil$ and $\lfloor r_a(c+d/2) \rfloor$). Further, the video frame and the spectrogram of the audio signal are fed into an audio-visual network corresponding to the current temporal segment. It is worth noting that $N$ independent audio-visual networks are needed to build the entire emotion recognition architecture.

\begin{figure*}[!t]
  \centering
  \includegraphics[scale = 0.35]{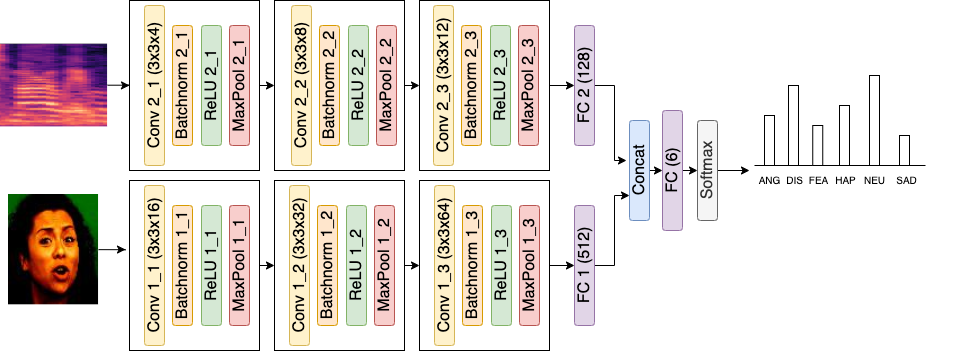}
  \caption[]{Our proposed audio-visual network used to retrieve emotion class probabilities for each temporal segment of the analysed video.}
  \label{fig:net}
\end{figure*}

Following a similar approach to the method that has been recently proposed by the authors in \cite{ristea2019emotion}, the core audio-visual network is composed of a sequence of convolutional blocks, which extracts the audio and visual features. After the concatenation of the audio and video feature vectors, the resulting feature vector is considered as input for a Fully Connected (FC) layer, followed by a SoftMax activation function which yields the class probabilities for the considered emotion categories. The solution proposed in \cite{ristea2019emotion} achieved approximately the human rating performance accuracy at low computational costs. However, in order to accelerate the training process and to increase the stability of the core audio-visual network, we inserted a Batch Normalization layer after each convolutional layer \cite{ioffe2015batch}. The core audio-visual network architecture, which processes asynchronous multimodal information, is shown in Fig.~\ref{fig:net}.

After aggregating the emotion class probabilities for all the temporal segments composing the video, the class label for the entire video is the one for which the maximum score is achieved. As shown in Fig.~\ref{fig:tbn}, the final score is obtaining by summing up the emotion class probabilities over all the temporal segments.

We mention that the frames of the original videos are pre-processed using the MTCNN algorithm \cite{7553523}, whose aim is to perform face detection and to remove the unnecessary information (i.e., background) with respect to the emotion recognition task. 


\section{Database}
Over the past years, several databases for the emotion recognition task have been proposed and the research in the affective computing domain focused on mixing different sources of information to achieve better performance. The  CREMA-D  multimodal database  was  published  in  2015 \cite{cao2014crema} and  contains 7442 clips of 91 actors (48 male and 43 female) with different ethnic backgrounds. The actors were asked to convey particular emotions while producing, with different intonations, 12 particular sentences that evoke the target emotions. Six labels have been used to discriminate among different emotion categories (i.e., neutral,  happy,  anger,  disgust,  fear,  sad)  with  four different intensity levels (i.e., low, medium, high, unspecified). The labels corresponding  to  each  recording  were  collected  using  crowd-sourcing. More  precisely,  2443  participants  were  asked  to  label  the perceived emotion and its intensity. The human accuracy achieved for this task was, on average, $63.6\%$. It is worth mentioning that human training was achieved through  participants' previous experience.



\section{Experiments}

\begin{figure}[!t]
\centering
\subfloat[Loss]{\includegraphics[width=0.37\textwidth]{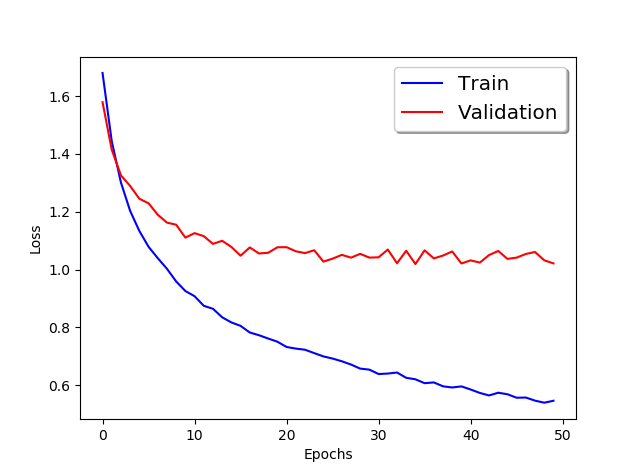}%
\label{fig: loss}}

\subfloat[Accuracy]{\includegraphics[width=0.37\textwidth]{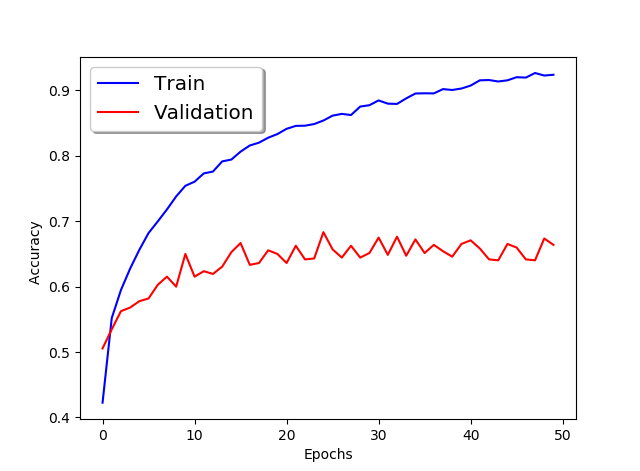}%
\label{fig: accuracy}}
\caption{Performance over the training and validation sets.}
\label{fig: loss + accuracy}
\end{figure}

\begin{figure}[!t]
  \centering
  \includegraphics[scale = 0.4]{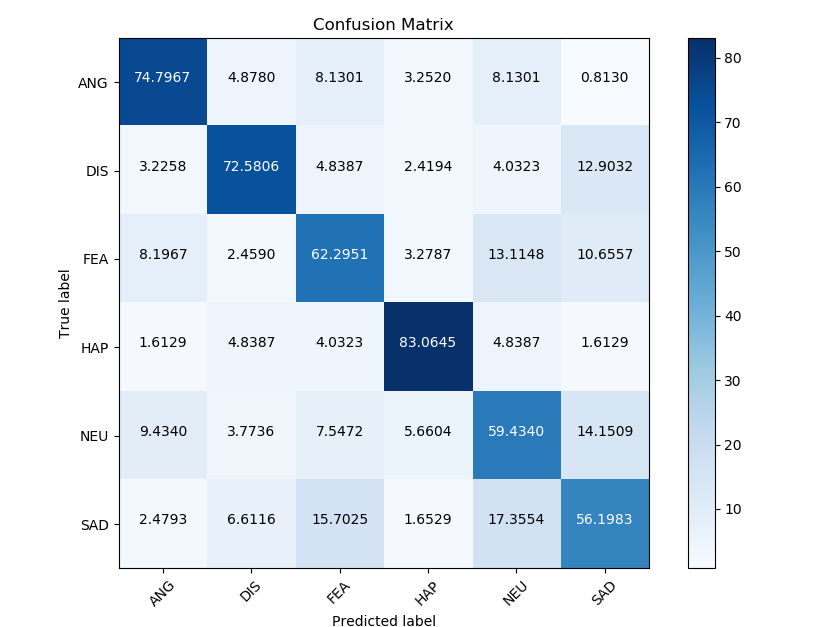}
  \caption[]{Confusion matrix for best performance.}
  \label{fig: confusion matrix}
\end{figure}

\begin{figure}[!t]
  \centering
  \includegraphics[scale=0.45]{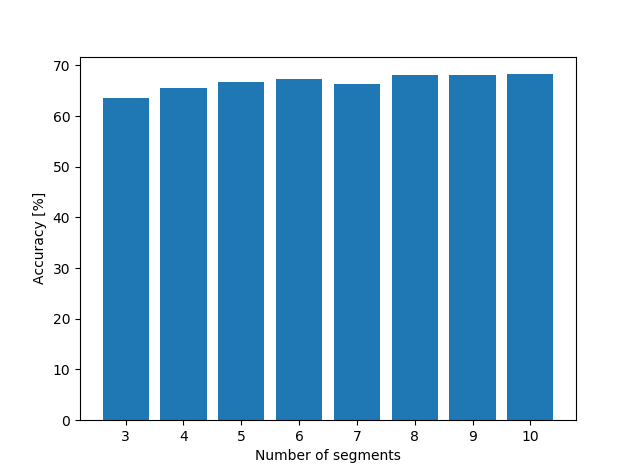}
  \caption[]{Variation of accuracy values with respect to the number of segments considered in the temporal aggregation.}
  \label{fig: Accuracy vs segments}
\end{figure}

In this section, the results achieved on the CREMA-D data set, as well as the experimental setup used in our approach, are provided. For all our experiments, we use an user-independent 10-fold cross validation technique to split the dataset into train and validation subsets. We divide the dataset into 10 different folds (i.e., none of the actors is introduced in more than one fold for generalization reasons) and compute the mean accuracy over all the folds.

Various evaluation metrics are used to assess the performance of the proposed solution for emotion recognition, namely mean overall accuracy, loss values and confusion matrix. These performance measures are shown in Fig.~\ref{fig: loss + accuracy} and Fig.~\ref{fig: confusion matrix} and are achieved for a temporal aggregation of 10 segments extracted from the video. As shown in Fig.~\ref{fig: Accuracy vs segments}, the overall accuracy increases with the number of segments considered. However, the time required for training a model and the inference time increase almost linearly with the number of segments, i.e., from 3 hours for a model with 3 segments to approximately 5.5 hours for a model with 10 segments. Moreover, the accuracy does not increase substantially for more than 8 segments. 

Furthermore, the overall accuracy is compared with the performances achieved by other methods from the literature (e.g., CNN-based approach \cite{ristea2019emotion}, RMA \cite{beard-etal-2018-multi}). It is worth mentioning that the experiments using the approach proposed in \cite{ristea2019emotion} followed the same 10-fold cross validation technique. 

The input frames were cut at $224 \times 224$ pixels around the detected faces \cite{7553523}, whereas the spectrograms were resized to $192 \times 120$. The length of the audio signal $d$ was set to 1.28 seconds, whereas the offset $b$ was set to 0.01 seconds. In order to train the models, we used the cross entropy loss and the stochastic gradient descent optimization method with 0.9 momentum. For each epoch, the learning rate was initially set to 1e-3 and decayed by a factor of 10 every 50 training steps. The batch size was set to 16. 

Our solution outperformed the baseline approach proposed in \cite{ristea2019emotion} by $ 12.6 \%$, human accuracy rating by $4.8 \%$ and also methods based on recurrent multi-attention \cite{beard-etal-2018-multi}.  
The proposed method of early combining features and temporal aggregation of partial results leads to a better performance, which suggests that combining information coming from different sources in an asynchronous manner proves to be beneficial in the process of emotion understanding.

We mention that all the experiments were conducted over an Intel Xeon E5-1680v3, 8 cores @3.2 GHz, equipped with NVIDIA Quadro M4000 GPU with 8 GB RAM.

\begin{table}[t]
\caption{Average accuracy rate on CREMA-D} 
\label{tabel: average acc}
\begin{center}
\begin{tabular}{ |c | c|} 
\hline
\textbf{Method} & \textbf{Accuracy [\%]} \\ 
\hline
Human accuracy \cite{cao2014crema} & 63.6 \\ 
CNN-based approach \cite{ristea2019emotion} & 55.8 \\
RMA \cite{beard-etal-2018-multi} & 65.0 \\
MERML \cite{8935376} & 66.5 \\ 
Proposed method & \textbf{68.4} \\
\hline
\end{tabular}
\end{center}
\end{table}

\section{Conclusion}
In this paper, we proposed a new method to incorporate multimodal information to discriminate among categories of emotions. The methodology benefits from combining the audio and visual information in an asynchronous manner which allows a certain degree of flexibility between the analysis of the two modalities. Using a simple audio-visual neural network as core architecture for predicting the emotional states, the temporal aggregation of the multimodal information from various segments leads to a substantial increase in the accuracy of the recognition system and outperforms other approaches from the literature.

\bibliographystyle{IEEEtran}
\bibliography{references}


\end{document}